\documentclass{article} 
\usepackage{iclr2023_conference,times}

\usepackage{amsmath,amsfonts,bm}

\def\eqref#1{equation~\ref{#1}}

\def\1{\bm{1}}

\DeclareMathAlphabet{\mathsfit}{\encodingdefault}{\sfdefault}{m}{sl}
\SetMathAlphabet{\mathsfit}{bold}{\encodingdefault}{\sfdefault}{bx}{n}

\usepackage{graphicx}
\usepackage{hyperref}
\usepackage{booktabs}
\usepackage{url}
\usepackage{multirow}
\usepackage{algorithm}
\usepackage{algorithmic}
\usepackage[T1]{fontenc}
\usepackage{color, colortbl}
\hypersetup{
    colorlinks=true,
    linkcolor=red,
    citecolor=cyan,
    filecolor=magenta,      
    urlcolor=cyan,
    }
\definecolor{Gray}{gray}{0.9}

\title{Class Prototype-based Cleaner for Label Noise Learning}

\author{Jingjia Huang \thanks{Corresponding author.} \\
ByteDance Inc\\
Beijing, China \\
\texttt{huangjingjia@bytedance.com} \\
\And
Yuanqi Chen\\
School of Electronic and Computer Engineering \\
Peking University Shenzhen Graduate School \\
Shenzhen, China \\
\texttt{cyq373@pku.edu.cn @pku.edu.cn} \\
\AND
Jiashi Feng \& Xinglong Wu \\
ByteDance Inc\\
Beijing, China \\
\texttt{\{jshfeng,wuxinglong\}@bytedance.com} \\
}

\iclrfinalcopy 
\begin{document}

\maketitle

\begin{abstract}
Semi-supervised learning based methods are current SOTA solutions to the noisy-label learning problem, which rely on learning an unsupervised label cleaner first to divide the training samples into a  labeled set for clean data and an unlabeled set for noise data. 
Typically, the cleaner is   obtained  via fitting a mixture model to  the distribution of  per-sample training losses.  However, the modeling procedure is  \emph{class agnostic} and  assumes the loss distributions of clean and noise samples are the same across different classes. Unfortunately, in practice, such an assumption does not always hold due to the varying learning difficulty of different classes, thus leading to sub-optimal   label noise partition criteria. In this work, we reveal this long-ignored problem and propose a simple yet effective solution, named  \textbf{C}lass \textbf{P}rototype-based label noise \textbf{C}leaner (\textbf{CPC}). Unlike previous works treating all the classes equally, CPC fully considers loss distribution heterogeneity and applies  class-aware modulation to partition the clean and noise data. CPC takes advantage of loss distribution modeling and intra-class consistency regularization in feature space simultaneously  and thus can  better distinguish clean and noise labels. We theoretically justify the effectiveness of our method by explaining it from the Expectation-Maximization (EM) framework. Extensive experiments are conducted on the noisy-label benchmarks CIFAR-10, CIFAR-100, Clothing1M and WebVision. The results show that CPC consistently brings about performance improvement across all benchmarks. Codes and pre-trained models will be released at \url{https://github.com/hjjpku/CPC.git}.
\end{abstract}

\section{Introduction}
Deep Neural Networks (DNNs) have brought about significant progress to the computer vision community over past few years. One key to its success is the availability of large amount of training data with proper annotations. However, label noise is very common in real-world applications. Without proper intervention, DNNs would be easily misled by the label noise and yield  poor performance. 

In order to improve the performance of DNNs when learning with noise labels, various methods have been developed \citep{liu2020early, Li2020DivideMixLW, reed2014training, nishi2021augmentation}. Among them, semi-supervised learning based methods \citep{nishi2021augmentation,Li2020DivideMixLW} achieve the most competitive results. The semi-supervised   learning methods  follow a two-stage   pipeline. They first model the  loss distribution of training samples to construct a noise cleaner based on the ``small-loss prior" \citep{han2020survey}, which says in the early stage of training, samples with smaller cross-entropy losses are more likely to have clean labels. The prior is widely adopted and demonstrated to be highly effective in practice \citep{han2020survey}. Given the noise cleaner, the training samples are divided into a labeled clean set and an unlabeled noise set. Then, semi-supervised learning strategies like MixMatch \citep{berthelot2019mixmatch} are employed to train DNNs on the divided dataset. 

The key to their performance lies in the accuracy of the label-noise cleaner \citep{cordeiro2022longremix}. Usually, a single Gaussian Mixture Model (GMM) \citep{Li2020DivideMixLW} is used  to model the   loss distribution of all the training samples across different categories. However, this modeling procedure is class-agnostic, which assumes a DNN model has the same learning speed to fit the training samples in different categories, thus the same loss value on samples in different categories can reflect the same degree of noise likelihood.

\begin{figure}[t]
  \centering
  \includegraphics[width=1\columnwidth]{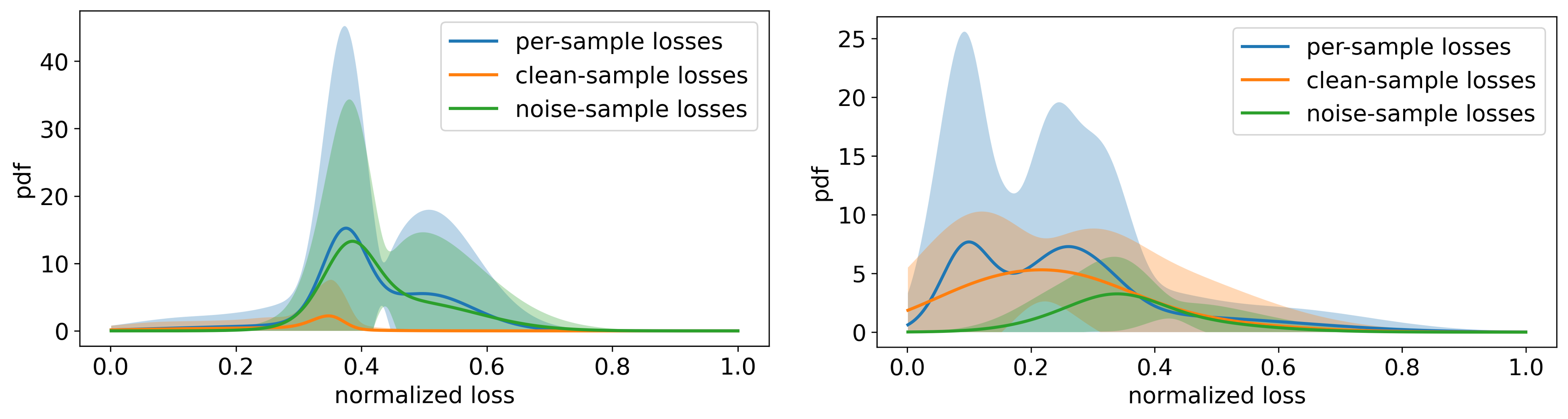}
  \caption{ Loss distribution of samples in CIFAR-100 with 90\% symmetric noise at epoch 30 (left)  and CIFAR-10 with 40\% asymmetric noise at epoch 10 (right), where the curves indicate mean probability density over all the categories while the shadow indicates the 95\% confidence interval. The loss distribution for each class deviates significantly from the average loss distribution.}
   \label{fig:motivation}
   \vspace{-1\baselineskip}
\end{figure}

Unfortunately, such assumption does not hold  in practise. 
In Fig.~\ref{fig:motivation}, 
we present the cross-entropy loss distribution of training samples at the end of DNNs warm-up period. We conduct Kolmogorov-Smirnov test \citep{massey1951kolmogorov} to quantify the loss distribution difference between the samples in each class and samples in the whole dataset. The results show that for 54\%  categories in CIFAR-100 under 90\% symmetric noise, the p-value is lower than 0.05\footnote{A p-value $<$ 0.05 suggests the probability that the class-wise loss distribution are the same with the global loss distribution is lower than 5\%.} for the hypothesis test that the probability distribution of clean samples in the class is the same with the probability distribution of clean samples in the whole dataset, while the number in the case of noise samples is 53\%. Therefore, the class-agnostic label noise cleaner, which establishes a overly rigid criterion shared by all the classes, would introduce more noise samples to the clean set while reject clean samples, and consequently get the model perform poorly. A straightforward remedy to the problem is to fit distinct GMMs to losses of samples in different classes respectively, yielding a class-aware GMM cleaner. Nevertheless, this class-aware modeling strategy implicitly assumes that label noise is existed in every class. In the case of asymmetric noise \emph{e.g.,} CIFAR10-asym40\%, where samples in parts of classes are clean, such a naive strategy would classify most of hard samples in the clean classes as noise, and results in negative affect on model training.


Considering that images in the same category should share similar visual representations, the similarity between a sample and the cluster center (\emph{e.g.}, class prototype) of its labeled class is helpful for recognizing label noise. In this paper, we propose a simple \textbf{C}lass \textbf{P}rototype-based label noise \textbf{C}leaner (\textbf{CPC}) to apply class-aware modulation to the partitioning of clean and noise data, which takes advantage of intra-class consistency regularization in feature space and loss distribution modeling, simultaneously. CPC learns embedding for each class, \emph{i.e.,} class prototypes, via intra-class consistency regularization, which urges samples in the same class to gather around the corresponding class prototype while pushes samples not belonging to the class away.
Unlike the aforementioned naive class-aware GMM cleaner, CPC apply class-aware modulation to label noise partitioning via representation similarity measuring without assuming that label noise is existed in every class, which is more general for different label noise scenarios. Meanwhile, CPC leverages the ``small-loss prior'' to provide stronger and more robust supervision signals to facilitate the learning of prototypes. 

We plug CPC to the popular DivideMix\citep{Li2020DivideMixLW} framework, which iterates between label noise partitioning and DNNs optimization. With the stronger label noise cleaner in the first stage, DNNs can be trained better in the second stage, which would further improve the learning of class prototypes. We theoretically justify the procedure from Expectation-Maximization algorithm perspective, which guarantees the efficacy of the method. We conduct extensive experiments on multiple noisy-label benchmarks, including CIFAR-10, CIFAR-100, Clothing1M and WebVision. The results clearly show that CPC effectively improves accuracy of label-noise partition, and brings about consistently performance improvement across all noise levels and benchmarks. 

The contribution of our work lie in three folds: (1) We reveal the long-ignored problem of \emph{class-agnostic} loss distribution modeling that widely existed in label noise learning, and propose a simple yet effective solution, named  Class Prototype-based label noise Cleaner (CPC); (2) CPC takes advantage of loss distribution modeling and intra-class consistency regularization in feature space simultaneously, which can  better distinguish clean and noise labels; (3) Extensive experimental results show that our method achieves competitive performance compared to current SOTAs.

\section{Related Work}
\label{sec:related_work}
Recent advances in robust learning with noisy labels can be roughly divided into three groups.
(a) \textbf{Label correction methods} aim to translate wrong labels into correct ones.
Early studies rely on an auxiliary set with clean samples
for clean label inference \citep{Xiao2015LearningFM,Vahdat2017TowardRA,Li2017LearningFN,Lee2018CleanNetTL}.
Recent efforts focus on performing label correction procedures without supervision regarding clean or noise labels.
\citep{Yi2019ProbabilisticEN,Tanaka2018JointOF} propose to jointly optimize labels during learning model parameters.
\citet{li2020mopro} propose to correct corrupted labels via learning class prototypes and utilize the pseudo-label generated by measuring the similarity between prototypes and samples to train model. \citet{wu2021ngc} and \citet{li2021learning} introduce neighbouring information in feature space  to correct noise label, and propose a graph-based method and a class prototype-based method, respectively. 
(b) \textbf{Sample selection methods}
select potential clean samples for training to eliminate the effect of noise labels on learning the true data distribution.
\citep{Han2018CoteachingRT,Jiang2018MentorNetLD,Jiang2020BeyondSN,Yu2019HowDD} involve training two DNNs simultaneously and focus on the samples that are probably to be correctly labeled.
(c) \textbf{Semi-supervised learning methods} conceal noise labels and treat these samples as unlabeled data \citep{Ding2018AST}. DivideMix \citep{Li2020DivideMixLW} is a typical algorithm among these works, which compromises an unsupervised label noise cleaner that divides the training data to a labeled clean set and an unlabeled noise set, followed by semi-supervised learning that minimize the empirical vicinal risk of the model. Inspired by DivideMix, a series of methods \citep{cordeiro2022longremix,nishi2021augmentation,Cordeiro2021PropMixHS} are proposed, which achieve SOTA performance. However, all these methods rely on  the class-agnostic loss distribution modeling to achieve the label noise cleaner, which hinders the performance of the model. The class-agnostic loss distribution modeling implicitly assumes a DNN model has the same learning speed to memory training samples in different categories. However, in reality, the memorization speed are actually different and will cause the the problem of under learning in hard classes as revealed by \citet{wang2019symmetric}. In this paper, we focuses on another problem, \emph{i.e.,} class agnostic loss distribution modeling problem caused by the issue in the context of label noise cleaner. In our method, we propose the simple yet effective class prototype-based label noise cleaner to solve the problem. Besides, compared to previous prototype-based label noise learning methods \citep{li2020mopro,li2021learning}, our method are different from them in two folds: (1) we utilize prototypes as label noise cleaner to effectively improve the semi-supervised learning based methods; (2) CPC takes advantage of both loss distribution modeling and intra-class consistency regularization in feature space simultaneously which learns better prototypes.

\begin{figure}[t]
  \center
  \includegraphics[width=0.9\columnwidth]{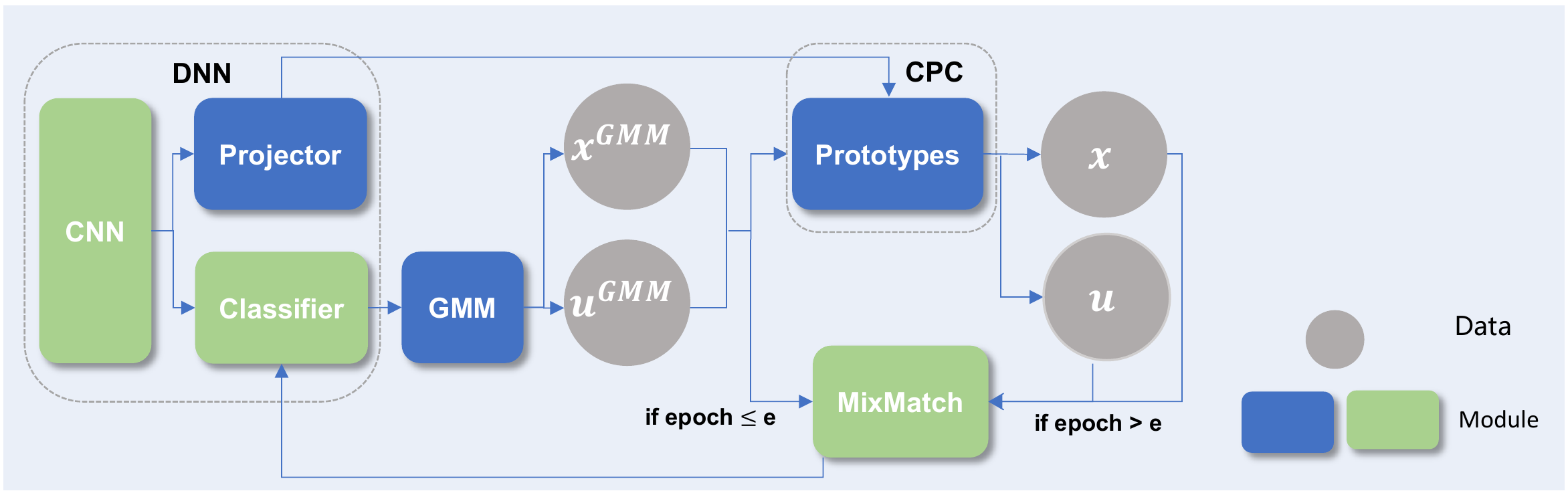}
  \caption{Illustration of the training pipeline in a single epoch. Blue modules are utilized in the first stage, where we update the prototypes in CPC and partition the training data.
   Green modules are utilized in the second stage, where the DNN model is trained based on the partitioned data.}
  \vspace{-1\baselineskip}
   \label{fig:pipeline}
\end{figure}

\section{Preliminary}\label{sec:preliminary}
In label noise learning, given a training set $D=(X, Y)=\{(x_i, y_i)\}_{i=1}^{N}$, where $x_i$ is an image and $y_i \in \{1,2,...,K\}$ is the annotated label over $K$ classes,  the label $y_i$ could differ from the unknown true label $\hat{y_i}$. 
In this paper, we follow the popular label noise learning framework DivideMix \citep{Li2020DivideMixLW}, which first warms up the model for a few epochs by training on all the data using the standard cross-entropy loss, and then trains the model by iterating a two-stage   pipeline. The pipeline comprises an unsupervised label cleaner $Q$ to divide training samples into a labeled set for clean data $\mathcal{X}$ and an unlabeled set for noise data $\mathcal{U}$, followed by a semi-supervised learning stage that trains the model to minimise the empirical vicinal risk (EVR) \citep{zhang2017mixup}:
\begin{equation} 
\ell_{EVR} = \frac{1}{|\mathcal{X'}|} \sum_{\mathcal{X'}} \ell_{\mathcal{X'}}(p(\tilde{y'_i}|x'_i), y'_i) + \frac{\lambda}{|\mathcal{U'}|} \sum_{\mathcal{U'}} \ell_{\mathcal{U'}}(p(\tilde{y'_i}|x'_i), y'_i),
\label{loss_EVR}
\end{equation}
where $\mathcal{X'}$ and $\mathcal{U'}$ indicate MixMatch \citep{berthelot2019mixmatch} augmented clean and noise set.  $l_{\mathcal{X'}}$ and $l_{\mathcal{U'}}$ denote the losses for samples in set $\mathcal{X'}$ and $\mathcal{U'}$, which are weighted by $\lambda$.  $p(\tilde{y'_i}|x'_i)$ is the softmax output of DNNs, where $\tilde{y'_i}$ is the predicted label. For more details about EVR, please refer to the appendix~\ref{app:evr}. 

In \citet{Li2020DivideMixLW}, the unsupervised label cleaner is operated under the ``small-loss prior", which is widely adopted and demonstrated to be highly effective \citep{han2020survey}.
 The prior assumes that in the early stage of training, samples with smaller cross-entropy losses are more likely to have clean labels. The well known insight behind the ``small-loss prior" is that DNNs tend to learn simple patterns first before fitting label noise \citep{arpit2017closer}. Given a training sample $x_i$ and the softmax output $p(\tilde{y_i}|x_i)$ of DNNs, where $\tilde{y_i}$ is the predicted label, the cross-entropy loss $l(p(\tilde{y_i}|x_i), y_i)$ reflects how well the model fits the training sample. 

To achieve the unsupervised label cleaner $Q$, a two-component Gaussian Mixture Model (GMM) is employed to fit the loss distribution of all training samples, \emph{i.e.,} $\ell(p(\tilde{y_i} |x_i), y_i) \sim  \phi_{0}\mathcal{N}(\mu_{0},\sigma_{0})+\phi_{1} \mathcal{N}(\mu_{1},\sigma_{1})$, where $\mu_{0} < \mu_{1}$, and $\phi$ is a mixing coefficient.
The component with smaller mean represents the distribution of clean samples and the other one is for noise samples.
We use $z_i \in \{0,1\}$ indicates the data is clean or not. Then, $q(z_i=0)$ represents the clean probability of $x_i$, which is the posterior probability of its loss belonging to the clean component.  The label cleaner is shared by training samples across different classes, which is actually \textit{class-agnostic}.
A hypothesis implicitly accompanying this loss distribution modeling method is ignored by current works, which assumes the loss distributions of clean and noise samples are consistent across different categories.
 Unfortunately, as illustrated in Fig.\ref{fig:motivation}, the  hypothesis dose not hold in practise. In this paper, we propose the class prototype-based label noise cleaner which applies class-aware modulation to the partitioning of clean and noise data and improves label noise learning.

\section{Methodology}

\subsection{Overview} \label{overview}
Our method follows the two-stage label noise learning framework DivideMix \citep{Li2020DivideMixLW} and improves the framework with the proposed CPC. CPC comprises class prototypes $C=\{c_k \in \mathbb{R}^{1 \times d}|k=1,2,...,K\}$, where $c_k$ indicates the prototype of $k$-th class and $d$ is the dimension of prototype embedding.
Our DNN model consists of a CNN backbone, a classifier head and a projection layer. The backbone maps an image input $x_i$ to a feature vector $v_{i} \in \mathbb{R}^{1 \times D}$. The classifier takes $v_{i}$ as input and outputs class prediction $p(\tilde{y_i}|x_i)$. The projection layer serves to project the high dimension feature $v_{i}$ to a low-dimensional embedding $v'_{i} \in \mathbb{R}^{1 \times d}$, where $d < D$. 

As shown in Fig.~\ref{fig:pipeline}, we update the DNN as well as the CPC by iterating a two-stage training pipeline in every epoch. In the first stage, we update CPC as well as the projector in DNN, and utilize the updated CPC to partition label noise. We first calculate the cross-entropy loss of every training sample and fits a GMM to the losses. We utilize the GMM as a label noise cleaner to get a labeled clean set $\mathcal{X}^{GMM}$ and a unlabeled noise set $\mathcal{U}^{GMM}$. The data partition $\mathcal{X}^{GMM}$ and $\mathcal{U}^{GMM}$ are utilized to update the prototypes in CPC and parameters in the projector. Note that we cut off the gradient back-propagation from the projector to the CNN backbone. Then, the updated CPC is employed to re-divide the training data into another two set $\mathcal{X}$ and $\mathcal{U}$. In the second stage, we train DNN model to minimise the EVR in Eq.~(\ref{loss_EVR}) with data partitioned by the cleaner. In the first $e$ epochs, we wait CPC to warm up, and minimise the EVR of DNNs based on training data partitioned by the GMM cleaner. After the $e$-th epoch, the label noise estimation results of CPC, \emph{i.e.,} $\mathcal{X}$ and $\mathcal{U}$ are employed to train DNNs, while the estimation results of GMM cleaner are only used to update prototypes in CPC. In inference, we utilize DNN classifier for image recognition, directly. In \ref{app:alg}, we further delineate the full framework.

\subsection{Class Prototype-based Label Noise Cleaner}
In order to apply class-aware modulation to the label noise partitioning, we propose to learn an embedding space where samples from the same class are aligned with their class prototypes, and leverage the prototypes to recognize noise labels. The prototypes are typically learnt with intra-class consistency regularization, which urges samples in the same class to align with the corresponding class prototype while keeping samples not belonging to the class away. Previous methods \citep{wang2022pico,li2020mopro} apply the intra-class consistency regularization to prototype learning via  unsupervised contrastive objectives, \emph{e.g.,} prototypical contrastive objective \citep{li2020prototypical}, where the unsupervised training labels are typically determined by the similarity between samples and prototypes. The 
accuracy of the training labels are highly depends on the quality of representation learnt by the CNN encoder, which would be too low to effectively update the prototypes, especially in the early stage of training. In contrast, we empirically find that the GMM cleaner, which is operated under the well evaluated ``small-loss prior'', are not as sensitive as the prototypes to the representation quality, and can provide more robust and accurate training labels. 

Therefore, we propose to take samples in clean set $\mathcal{X}^{\mathrm{GMM}}$ as positive samples and those in noise set $\mathcal{U}^{\mathrm{GMM}}$ as negative samples to update prototypes. Specifically, given the feature embedding $v'_i$ of a sample $x_i$ from $\mathcal{X}^{\mathrm{GMM}}$, we update prototypes $C$ as well as the parameters of the projector to maximize the score $q(z_i=0)$ between $c_{k=y_i}$ and $v'_i$, and minimize the score between $c_{k \neq y_i}$ and $v'_i$ via minimize $L_{\mathcal{X}^{\mathrm{GMM}}}$: 
\begin{equation}
\begin{aligned}
L_{\mathcal{X}^{\mathrm{GMM}}} &= - \frac{1}{|\mathcal{X}^{GMM}|} \sum_{\mathcal{X}^{\mathrm{GMM}}} \sum_{k=1}^{K} \ell_{k}(v'_i,y_i), ~\text{ where } \\
\ell_{k}(v'_i,y_i) &= \left\{
\begin{aligned}
& \log(\mathrm{sigmoid}(v'_i c_k^\top)), &k=y_i,\\
& \lambda^{neg} \log(1-\mathrm{sigmoid}( v'_i  c_k^\top)), & k \neq y_i,
\end{aligned}
\right.
\end{aligned}
\end{equation}
where $\lambda^{neg} = \frac{1}{K}$ weights the losses between positive pair and negative pairs to avoid under-fitting the positive samples. Given $v'_i$ of a sample $x_i$ from $ \mathcal{U}^{\mathrm{GMM}}$, we update prototypes $c_k$ as well as the parameters of the projector to minimize the score $q(z_i=0)$ between $ c_{k = y_i}$ and $ v'_i$ via minimizing $L_{\mathcal{U}^{\mathrm{GMM}}}$:
\begin{equation}
L_{\mathcal{U}^{\mathrm{GMM}}} = - \frac{1}{|\mathcal{U}^{GMM}|} \sum_{\mathcal{U}^{\mathrm{GMM}}} \log(1-\mathrm{sigmoid}( v'_i  c_k^\top)), \text{ where}~ k=y_i.
\end{equation}
At last, for noise samples in $\mathcal{U}^{\mathrm{GMM}}$ with high classification confidence, the samples are more likely to belong to the class predicted by DNNs, which is potentially valuable to the update of prototypes. 
Therefore, we collect such training samples $\mathcal{X}^{P}$ from $\mathcal{U}^{\mathrm{GMM}}$ taking the averaged classification confidence of samples in $\mathcal{X}^{\mathrm{GMM}}$ as the threshold. Specifically, given a sample in $\mathcal{U}^{\mathrm{GMM}}$ with the label predicted by DNNs $k=\max_k(p(\tilde{y_i}|x_i))$, the sample is collected into $\mathcal{X}^{P}$ if   $p(\tilde{y_i}|x_i)_k > average(\{p(\tilde{y_i}|x_i)_k| (x_j,y_j|y_j=k) \in \mathcal{X}^{\mathrm{GMM}}\})$. Then, we update the prototypes and projectors to minimize $L_{(\mathcal{X}^{P})}$:
\begin{equation}
L_{\mathcal{X}^{P}} = - \frac{1}{|\mathcal{X}^{P}|} \sum_{\mathcal{X}^{P}} \log(\mathrm{sigmoid}( v'_i  c_k^\top)), \text{where}~ k=\max_k(p(\tilde{y_i}|x_i)).
\end{equation}
The overall empirical risk $L_{C}$ for prototypes and the projector is as follows:
\begin{equation} \label{fullloss}
L_{C} = L_{\mathcal{X}^{\mathrm{GMM}}} + L_{\mathcal{U}^{\mathrm{GMM}}} + \alpha L_{\mathcal{X}^{P}},
\end{equation}
where $\alpha$ is the weight scalar.

 CPC distinguishes a clean sample $(x_i,y_i)$ with the score $q(z_i=0)=\mathrm{sigmoid}(v'_i c_{k=y_i}^\top)$ and the threshold $\tau$. Samples with $q(z_i=0) > \tau$ are classified as clean, and otherwise as noise.


\subsection{Theoretical Justification on the Efficacy of CPC} \label{sec:em}
We provide theoretical justification on the efficacy of CPC from the perspective of Expectation-Maximization algorithm, which guarantees that though CPC does not follow the classical prototypical contrastive objective, it can  still learn meaningful prototypes and act as an effective cleaner. 

We consider training data with label noise $D=(X, Y)={(x_i, y_i)}_{i=1}^{N}$ as the observable data, and $Z \in  \{0,1\}^N$ as the latent variable, where $z_{i}=0$ \emph{iff} $(x_i, y_i)$ is clean (\emph{i.e.,} $y_i = \hat{y_i}$). The prototypes $C$ in the cleaner are taken as parameters expected to be updated. Then, the negative log likelihood for $D$ given $C$ is as follows:
\begin{equation}
NLL(D|C)  = - \sum_{ D} \log \sum_{z_i \in \{0,1\}} p(x_i,y_i,z_i|C) = - \sum_{D} \log \sum_{z_i \in \{0,1\}} q(z_i) \frac{p(x_i,y_i,z_i|C)}{q(z_i)},
\end{equation}
where $q(z_i) = p(z_i|x_i,y_i,C)$. According to the Bayes theorem and Jensen's inequality , we have
\begin{equation}
\begin{aligned}
NLL(D|C) & =  - \sum_{ D} \log \sum_{z_i \in \{0,1\}} q(z_i) p(x_i,y_i|C),\\
& \leq - \sum_{ D} \sum_{z_i \in \{0,1\}} q(z_i) \log   p(x_i,y_i|C)\\
& = - \sum_{ D} \sum_{z_i \in \{0,1\}} q(z_i) \log   p(y_i|C,x_i) + const,
\end{aligned}
\end{equation}
where $- \sum_{ D} \sum_{z_i \in \{0,1\}} q(z_i) \log   p(y_i|C,x_i)$ is the upper bound of $NLL(D|C)$. 
Typically, we can adopt the EM algorithm to find the prototypes $C$ that minimize the upper bound by iterating:

\textbf{E-step}: Compute a new estimate of $q(z_i)$ (\emph{i.e.,} clean or noise) according to prototypes $C^{old}$ from the last iteration:
    \begin{equation} \label{estep}
        q(z_i) = p(z_i|x_i,y_i,C^{old}).
    \end{equation}
    \textbf{M-step}: Find the prototypes $C$ that minimizes the bound:
    \begin{equation} \label{mstep}
        C^{new} =  \mathop{\arg\min}_{C} -   \sum_{ D} \sum_{z_i \in \{0,1\}} q(z_i) \log   p(y_i|C,x_i)
    \end{equation}
In our method, in order to introduce the ``small-loss prior'' to provide stronger and more robust supervision signals to the learning of CPC, in the \textbf{E-step}, we estimate the distribution of clean or noise of samples, denoted as $q(z'_i)$, via the GMM cleaner instead of $q(z_i)$ in Eq.~(\ref{estep}). And consequently, we replace the $q(z_i)$ in  Eq.~(\ref{mstep}) to $q(z'_i)$ and find the prototype $C$ minimize the bound. Next, we provide the justification that the EM algorithm still work by proving that $q(z'_i)$ can be considered as an approximation to $q(z_i)$ in our framework.


In our method, $q (z'_i)=p(z'_i|l(p(\tilde{y}_{i}|x_{i}), y_i))$, where $\tilde{y}_i \sim p(\tilde{ y_{i}}|x_{i},\theta)$, 
which is the label predicted by the DNN parameterized by $\theta$.
As introduced in section~\ref{overview}, in the first stage of each epoch, the CPC's estimation results $z_i \sim q(z_i)$ are utilized to divide training samples into a labeled set for clean  data $\mathcal{X}=\{(x_i,y_i) | z_i = 0\}$ and an unlabeled set for noise data $\mathcal{U}=\{(x_i,y_i) | z_i = 1\}$.
Then the parameters of DNNs, which we denote as $\theta$, are optimized using Eq.~(\ref{loss_EVR}) in the second stage.
There exists an optimal $\theta^*$ with respected to $z_i$, with which the softmax output $p(\tilde{y}_i|x_i)$ of DNNs satisfies:
\begin{equation}
    \ell(p(\tilde{y}_{i}|x_{i}), y_i) =  0, \text{ if } z_i=0, \text {otherwise } 1,
\end{equation}
where $\ell(p(\tilde{y_i}|x_i), y_i)$ is the cross-entropy loss between the network prediction and the annotated label.
With these loss values,
the subsequent GMM cleaner can easily distinguish samples of $\mathcal{X}$ from samples of $\mathcal{U}$.
In other words,
under the optimal $\theta^*$,
the estimation of the GMM cleaner would be consistent with the partition of CPC, \emph{i.e.,} $z'_i = z_i$. In practice, in each epoch, we takes the $\theta$ optimized to minimize Eq.~(\ref{loss_EVR}) as an approximation to the optimal $\theta^*$ with respect to $z_i$, and consequently we can get $q(z'_i)$ as an approximation to $q(z_i)$. Therefore, we can see that with the ``small loss prior'' introduced into the prototype learning, the EM optimization procedure would still work, which guarantees CPC can learn meaningful prototypes and act as an effective cleaner.
In appendix~\ref{app:kl}, we further present more details and empirical results to demonstrate the approximation is hold in practice. 

\section{Experiments}\label{sec:exp}
\subsection{Datasets and Implementation Details}
\textbf{Datasets.} We evaluate our method on the following popular LNL benchmarks. For CIFAR-10 and CIFAR-100 \citep{krizhevsky2009learning}, we experiment with two types of synthetic noise: symmetric and asymmetric, which are injected into the datasets following the standard setup in \citep{Li2020DivideMixLW}. Clothing1M \citep{xiao2015learning} and WebVision1.0 \citep{li2017webvision} are two large-scale real-world label noise benchmarks. Clothing1M contains 1 million images in 14 categories acquired from online shopping websites, which is heavily imbalanced and most of the noise is asymmetric \citep{yi2019probabilistic}. WebVision1.0 contains 2.4 million images crawled from the web using the concepts in ImageNet-ILSVRC12 (ILSVRC12). Following convention, we compare with SOTAs on the first 50 classes of WebVision, as well as the performance after transferring to ILSVRC12.  

\textbf{Implementation details.} We plug the proposed CPC to the DivideMix \citep{Li2020DivideMixLW} framework. For Clothing1M and CIFAR-10 with asymmetric noise, we employ a single class-agnostic GMM for loss-distribution modeling. For other cases, we find that  class-aware GMMs would further improve the performance of CPC. Following DivideMix, we employ ResNet18 \citep{he2016identity} for CIFAR-10 and CIFAR-100, and utilize ImageNet pre-trained ResNet-50 for Clothing1M. Since previous works chose different backbones, \emph{e.g.}, Inception-resnet v2 \citep{szegedy2017inception} and ResNet-50, we adopt the weaker one, \emph{i.e.,} ResNet-50 according to \citep{DBLP:journals/corr/abs-2103-13646}, and train it from scratch for fair comparison. The threshold of CPC $\tau$ is set $0.5$ by default for all the datasets except for the extremely imbalanced Clothing1M where it is set to $0.3$. For CIFAR-10 and CIFAR-100, we train the models for 450 epochs. For the large-scale dataset Clothing1M and WebVision1.0, we train the model for 80 and 100 epochs, respectively. The warm-up periods of prototypes for all the datasets is set to the first 5\% epochs after network warm-up, except in CIFAR-100 with noise ratios larger than 80\% when set to 10\% of total epochs. For the other settings, we simply follow the standard set-up as in DivideMix. For more implementation details, please refer to the appendix \ref{app:conig} and codes in supplementary materials.
\begin{table}[h]
\vspace{-1\baselineskip}
\caption{Comparison with SOTAs on Real-world Benchmarks. Following GJS\citep{englesson2021generalized}, we run our method three times with different random seeds and report the mean and standard deviation of classification accuracy. $\lozenge$ indicates methods utilize ResNet50 for WebVision, while others utilize Inception-resnet v2. The best results are indicated with  boldface.}
\label{table-sota-real}
\small
\begin{center}
\begin{tabular}{c|cc|cc|c}
 & \multicolumn{2}{c|}{WebVision} & \multicolumn{2}{c|}{WebVision $\rightarrow$ ILSVRC12} & Clothing1M  \\
 & top1 & top5 & top1 & top5 &  \\ \midrule
ELR+ & 77.78 & 91.64 & 70.29 & 89.76 & 74.8 \\
DivideMix & 77.32 & 91.64 & 75.2 & 90.84 & 74.76 \\
DivideMix$^{\lozenge}$ & $76.3_{\pm 0.36}$ & $90.65_{\pm 0.16}$ & $74.42_{\pm 0.29}$ & $91.21_{\pm 0.12}$ & 74.76 \\
LongReMix & 78.92 & 92.32 & - & - & 74.38 \\
NGC & 79.16 & 91.84 & 74.44 & 91.04 & - \\
AugDMix & - & - & - & - & 75.11 \\
NCR$^{\lozenge}$ & \textbf{80.5} & - & - & - & 74.6 \\
GJS$^{\lozenge}$ & $79.28_{\pm 0.24}$ & $91.22_{\pm 0.3}$ & $75.5_{\pm 0.17}$ & $91.27_{\pm 0.26}$ & - \\ \midrule
Baseline$^{\lozenge}$ & $76.3_{\pm 0.36}$ & $90.65_{\pm 0.16}$ & $74.42_{\pm 0.29}$ & $91.21_{\pm 0.12}$ & $74.73_{\pm 0.02}$ \\
Ours$^{\lozenge}$ & $79.63_{\pm0.08}$ & \textbf{93.46$_{\pm 0.10}$} & \textbf{75.75$_{\pm0.14}$} & \textbf{93.49$_{\pm0.25}$} & \textbf{75.40$_{\pm0.10}$} 
\end{tabular}%
\end{center}
\vspace{-2\baselineskip}
\end{table}

\subsection{Comparison with State-of-the-art methods}
\textbf{Real-world noise benchmarks.} We evaluate our method on real-world large scale data sets, and compare our method with latest SOTA label noise learning methods, including DivideMix\citep{Li2020DivideMixLW}, LongReMix\citep{cordeiro2022longremix}, NGC\citep{wu2021ngc}, GJS\citep{englesson2021generalized}, ELR+\citep{liu2020early}, AugDMix\citep{nishi2021augmentation} and NCR\citep{huang2021learning}. For WebVision, we measure the top1 and top5 accuracy on WebVision validation set and ImageNet ILSVRC12 validation set. We take ResNet50-based DivideMix \citep{DBLP:journals/corr/abs-2103-13646} as baseline. As shown in Table~\ref{table-sota-real}, our CPC improves top1 and top5 accuracy over baseline model on WebVision by 3.33\% and 2.81\%, respectively. Our method achieves competitive performance on WebVision, and shows stronger transferable capability,  outperforming other competitors on the ILSVRC12 validation set significantly. For Clothing1M, we apply the strong augmentation strategy \citep{nishi2021augmentation} to DivideMix as our baseline, and rerun the method three times. Our method achieves 75.4\% accuracy on this challenging benchmark, outperforming all the other SOTAs. We also notice that though NCR achieves SOTA result on WebVision, it shows moderate performance compared to ELR+, DivideMix and AugDMix on Clothing1M  containing asymmetric noise with imbalanced data distribution. It reveals that our method could be more robust across different label noise scenarios.

\begin{table}[t]
\vspace{-1\baselineskip}
\caption{Comparison with SOTAs on CIFAR-10 and CIFAR-100. Following previous work \citep{ wu2021ngc}, we run our method three times with different random seeds and report the mean and standard deviation of classification accuracy. $\dagger$ indicates our baseline. * indicates semi-supervised learning based label noise learning methods. SOTA results are indicated with  boldface.}
\label{table-sota-cifar}
\small
\begin{center}
\begin{tabular}{c|cccc|c}
 & \multicolumn{4}{c|}{CIFAR-10 / CIFAR-100 (Sym)} & CIFAR-10 (Asym) \\
 & 20\% & 50\% & 80\% & 90\% & 40\% \\ \midrule
ELR+ & 94.9  / 76.3  & 93.9. / 72.0 & 90.9. / 57.2 & 74.5  / 30.9 & 88.9 \\
NCR & 95.2  / 76.6 & 94.3 / 72.5 & 91.6  / 58.0 & 75.1  / 30.8 & 90.7 \\
ProtoMix & 96.4  / \textbf{80.3} & 95.3 / 76.0 & 93.3  / 61.1 & 77.4  / 33.1 & 92.6 \\
DivideMix* & 96.1  / 77.3 & 94.6  / 74.5 & 93.2  / 60.2 & 76.0  / 31.5 & 93.4 \\
LongReMix* & 96.2  / 77.8 & 95.0  / 75.6 & 93.9  / 62.9 & 82.0  / 33.8 & 94.7 \\
AugDMix*$^{\textbf{$\dagger$}}$ & 96.3  / 79.5 & 95.6  / 77.2 & 93.6  / 66.4 & 91.9  / 41.2 & 94.6 \\ \midrule
\multirow{2}{*}{NGC} & 95.88 / 79.31 & 94.54 / 75.91 & 91.59 /62.7 & 80.46 / 29.76 & 90.55 \\
  & $_{\pm}0.13$ / $_{\pm}0.35$ & $_{\pm}0.35$ / $_{\pm}0.39$ & $_{\pm}0.31$ / $_{\pm}0.37$ & $_{\pm}1.97$ / $_{\pm}0.85$ & $_{\pm}0.29$  \\
\multirow{2}{*}{GJS} & 95.33 / 75.71 & - & 79.11 / 44.49 & - & 89.65 \\
 & $_{\pm}0.18$ / $_{\pm}0.25$ & - & $_{\pm}0.31$ / $_{\pm}0.53$ & - & $_{\pm}0.37$ \\
\multirow{2}{*}{Ours*} & \textbf{96.50} / 80.22 & \textbf{95.64} / \textbf{79.31} & \textbf{94.78} / \textbf{69.56} & \textbf{92.55} / \textbf{54.60} & \multicolumn{1}{c}{\textbf{94.73}} \\
 & $_{\pm}0.10$ / $_{\pm}0.21$  & $_{\pm}0.01$ / $_{\pm}0.13$   & $_{\pm}0.01$ / $_{\pm}0.34$ & $_{\pm}0.58$ / $_{\pm}0.24$ & $_{\pm}0.04$ 
\end{tabular}
\vspace{-2\baselineskip}
\end{center}
\end{table}

\textbf{Synthetic noise benchmarks.} We evaluate the performance of CPC on CIFAR-10 and CIFAR-100 datasets with symmetric label noise level ranging from 20\% to 90\% and asymmetric noise of rate 40\%. We take AugDMix as the baseline, and compare our method with latest SOTA methods, where DivideMix, LongReMix and Aug-DMix are semi-supervised learning based methods. Following NGC and GJS, we run our method three times with different random seeds and report the mean and standard deviation. For other methods, \emph{e.g.,} ProtoMix \citep{li2021learning}, we report the best results reported in their papers. As shown in Table~\ref{table-sota-cifar}, though with a baseline method as strong as AugDMix, our method brings about performance improvement across all noise levels as well as noise types consistently, and establishes new SOTAs on CIFAR-10 and CIFAR-100. Additionally, we notice that, under asymmetric noise set-up, semi-supervised learning based methods consistently outperform other methods that achieve SOTA results on WebVision benchmark,including NGC, GJS and NCR. The results reveal that semi-supervised learning based method could be more robust to asymmetric noise, while our method achieves SOTA performance among them.

\subsection{Analysis}
\begin{figure}[t]
  \centering
  \includegraphics[width=1\columnwidth]{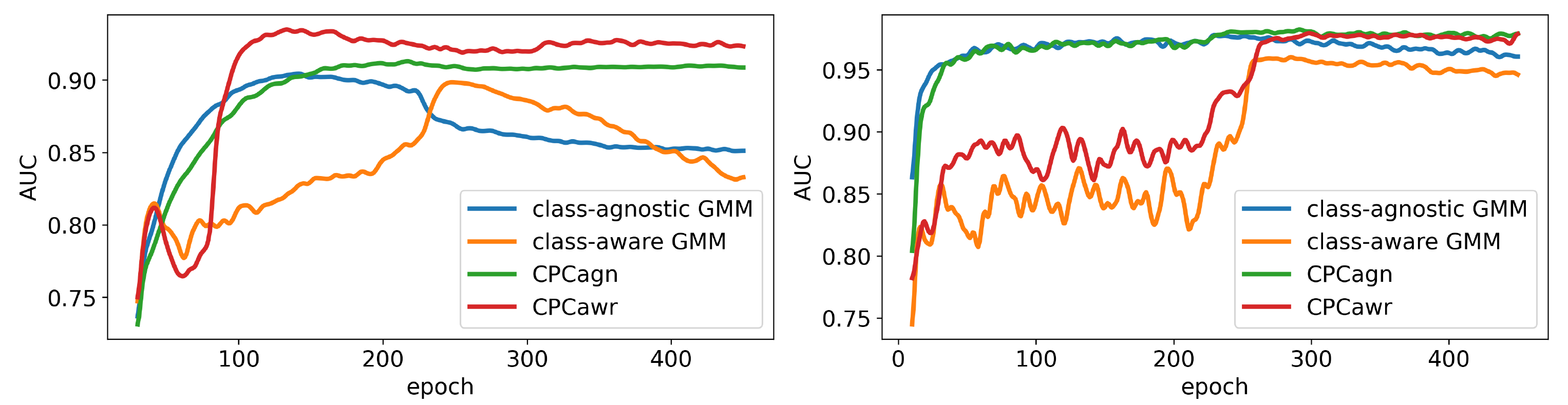}
    \vspace{-2\baselineskip}
  \caption{ AUC of different label noise cleaners with respect to the training period. Left is the results on CIFAR-100 under high symmetric noise ratio 0.9. Right is the results on CIFAR-10 under medium asymetric noise ratio 0.4.}
   \vspace{-1\baselineskip}
   \label{fig:auc}
\end{figure}

\textbf{Is CPC a better label noise cleaner?} We evaluate the performance of label noise cleaner under both symmetric and asymmetric label noise set-ups. For symmetric noise, we use CIFAR-100 with 90\% noise as benchmark to reveal the relationship between CPC and the significant performance improvement under this set-up. For asymmetric noise, we employ the most commonly adopted CIFAR-10-asym40\% as benchmark. The AUC of clean/noise binary classification results of a cleaner is calculated as the evaluation metric. We take the original class-agnostic GMM cleaner (GMM$_{agn}$) proposed in DivideMix as baseline, and compare it to our CPC and the aforementioned naive class-aware GMM cleaner (GMM$_{awr}$). Furthermore, we also implement another version of CPC that trained based on the class-aware GMM cleaner. To distinguish these two CPC, we denote the regular one trained based on conventional class-agnostic GMM cleaner as CPC$_{agn}$, and the other one as CPC$_{awr}$. As shown in Figure~\ref{fig:auc}, in both cases, the regular CPC$_{agn}$ outperforms the baseline GMM$_{agn}$ as well as GMM$_{awr}$, which demonstrates our class prototype-based method is the better label noise cleaner. As for the comparison between GMM$_{agn}$ and GMM$_{awr}$, we find that in the situation of high symmetric noise, though GMM$_{agn}$ shows better performance in the early stage of training, GMM$_{awr}$ outperforms it in the second half stage of training. In the case of asymmetric noise, GMM$_{awr}$, which tend to classify hard clean samples in clean categories as noise wrongly, consistently underperforms GMM$_{agn}$ across the whole training period. The results further prove that our class prototype-based method is the better choice for applying class-aware modulation to label noise cleaning, which is more robust across different noise types. Moreover, we find that in the case of asymmetric noise, CPC$_{agn}$ achieves higher AUC compared to GMM$_{agn}$, which shows our method can partially make up for the shortcomings of GMM$_{agn}$. In the case of symmetric noise, we find that GMM$_{agn}$ can further improve the performance of CPC, where CPC$_{awr}$ achieves the best performance among the four cleaners.

\textbf{How do different label noise cleaners affect label noise learning?} We plug different cleaners to DivideMix framework, and keep all the other training settings the same as described in the implementation details. As shown in Table~\ref{table-ablation}, the final performance of the model is consistent with the performance of the cleaner used. On CIFAR-100 with 90\% symmetric noise, performance improvement bought about by CPC$_{agn}$ are 7.68\%, while model with CPC$_{awr}$ outperforms the baseline method by 13.4\%. We also report the comparison results on large-scale WebVision dataset, where the performance of different models show the same trend of change as in CIFAR-100-sym90\%. As for the asymmetric noise situation, \emph{i.e.,} CIFAR-10-asym40\% and Clothing1M, model with CPC$_{agn}$, which has superior label noise partitioning capability as shown in Fig.\ref{fig:auc}, achieves best performance while CPC$_{awr}$ beat GMM$_{awr}$ in both cases.  The results demonstrate that CPC is helpful to train a better model in label noise learning.

\begin{table}[t]
\caption{The affect of different label noise cleaner to the final classification accuracy. The best results are indicated with  boldface.}
\label{table-ablation}
\small
\begin{center}
\begin{tabular}{c|c c c c}
 \multirow{2}{*}{Model}& \multicolumn{1}{c}{CIFAR-100} & {WebVision} &  CIFAR-10 & \multirow{2}{*}{Clothing1M} \\
 & sym-90\% & top1 & asym-40\%  &\\ \midrule
w/ GMM$_{agn}$ & 41.2 &76.32 & 94.6 & 74.73  \\
w/ GMM$_{awr}$ & 45.6 & 78.66 & 94.18 & 74.18 \\
w/ CPC$_{agn}$ & 48.88  & 79.4 & \textbf{94.73} & \textbf{75.4} \\
w/ CPC$_{awr}$ & \textbf{54.6}  & \textbf{79.63} & 94.29 & 74.36  
\end{tabular}%
\end{center}
\vspace{-2\baselineskip}
\end{table}

\textbf{Is the GMM cleaner beneficial to the learning of prototypes?}
In our method, we propose to leverage the GMM cleaner to facilitate the learning of prototypes via the ``small loss prior''. To validate the effectiveness of our method, we first compare the quality of prototypes learnt in CPC with prototypes learnt in another prototype-based label noise learning method MoPro \citep{li2020mopro}. We take WebVision as benchmark and utilize prototypes to classify test samples via measuring the similarity between samples and prototypes. The results show that, on the first 50 classes of WebVision, our prototype achieves a top1 accuracy of 78.44\%, while MoPro's accuracy is 72.23\%, which demonstrates that our method is able to learn better prototypes.
To further verify the contribution of the GMM cleaner, we remove the GMM cleaner and learn class prototypes in CPC via the typical prototypical contrastive objective as in MoPro. In experiments, we find that without the help of the GMM cleaner, the learnt prototypes generate less accurate data partition that further drawing back the overall training framework for DNNs, which proves the benefits of the GMM cleaner to our method. For more details and discussion, please refer to \ref{app:abl}.


\section{Conclusion}
In this paper, we reveal the long-ignored problem of \emph{class-agnostic} loss distribution modeling that widely existed in label noise learning, and propose a simple yet effective solution, named  Class Prototype-based label noise Cleaner (CPC). CPC takes advantage of loss distribution modeling and intra-class consistency regularization in feature space simultaneously, which can  better distinguish clean and noise labels. We  justify the effectiveness of our method by explaining it from the EM algorithm perspective theoretically and providing extensive empirical proves. The experimental results show that our method achieves competitive performance compared to current SOTAs.

\bibliography{iclr2023_conference}

\begin{thebibliography}{37}
\providecommand{\natexlab}[1]{#1}
\providecommand{\url}[1]{\texttt{#1}}
\expandafter\ifx\csname urlstyle\endcsname\relax
  \providecommand{\doi}[1]{doi: #1}\else
  \providecommand{\doi}{doi: \begingroup \urlstyle{rm}\Url}\fi

\bibitem[Arpit et~al.(2017)Arpit, Jastrz{\k{e}}bski, Ballas, Krueger, Bengio,
  Kanwal, Maharaj, Fischer, Courville, Bengio, et~al.]{arpit2017closer}
Devansh Arpit, Stanis{\l}aw Jastrz{\k{e}}bski, Nicolas Ballas, David Krueger,
  Emmanuel Bengio, Maxinder~S Kanwal, Tegan Maharaj, Asja Fischer, Aaron
  Courville, Yoshua Bengio, et~al.
\newblock A closer look at memorization in deep networks.
\newblock In \emph{International conference on machine learning}, pp.\
  233--242. PMLR, 2017.

\bibitem[Berthelot et~al.(2019)Berthelot, Carlini, Goodfellow, Papernot,
  Oliver, and Raffel]{berthelot2019mixmatch}
David Berthelot, Nicholas Carlini, Ian Goodfellow, Nicolas Papernot, Avital
  Oliver, and Colin~A Raffel.
\newblock Mixmatch: A holistic approach to semi-supervised learning.
\newblock \emph{Advances in neural information processing systems}, 32, 2019.

\bibitem[Cordeiro et~al.(2021)Cordeiro, Belagiannis, Reid, and
  Carneiro]{Cordeiro2021PropMixHS}
Filipe~R. Cordeiro, Vasileios Belagiannis, Ian~D. Reid, and G.~Carneiro.
\newblock Propmix: Hard sample filtering and proportional mixup for learning
  with noisy labels.
\newblock \emph{BMVC}, 2021.

\bibitem[Cordeiro et~al.(2022)Cordeiro, Sachdeva, Belagiannis, Reid, and
  Carneiro]{cordeiro2022longremix}
Filipe~R Cordeiro, Ragav Sachdeva, Vasileios Belagiannis, Ian Reid, and Gustavo
  Carneiro.
\newblock Longremix: Robust learning with high confidence samples in a noisy
  label environment.
\newblock \emph{Pattern Recognition}, pp.\  109013, 2022.

\bibitem[Ding et~al.(2018)Ding, Wang, Fan, and Gong]{Ding2018AST}
Yifan Ding, Liqiang Wang, Deliang Fan, and Boqing Gong.
\newblock A semi-supervised two-stage approach to learning from noisy labels.
\newblock \emph{2018 IEEE Winter Conference on Applications of Computer Vision
  (WACV)}, pp.\  1215--1224, 2018.

\bibitem[Englesson \& Azizpour(2021)Englesson and
  Azizpour]{englesson2021generalized}
Erik Englesson and Hossein Azizpour.
\newblock Generalized jensen-shannon divergence loss for learning with noisy
  labels.
\newblock \emph{Advances in Neural Information Processing Systems},
  34:\penalty0 30284--30297, 2021.

\bibitem[Han et~al.(2018)Han, Yao, Yu, Niu, Xu, Hu, Tsang, and
  Sugiyama]{Han2018CoteachingRT}
Bo~Han, Quanming Yao, Xingrui Yu, Gang Niu, Miao Xu, Weihua Hu, Ivor Wai-Hung
  Tsang, and Masashi Sugiyama.
\newblock Co-teaching: Robust training of deep neural networks with extremely
  noisy labels.
\newblock In \emph{NeurIPS}, 2018.

\bibitem[Han et~al.(2020)Han, Yao, Liu, Niu, Tsang, Kwok, and
  Sugiyama]{han2020survey}
Bo~Han, Quanming Yao, Tongliang Liu, Gang Niu, Ivor~W Tsang, James~T Kwok, and
  Masashi Sugiyama.
\newblock A survey of label-noise representation learning: Past, present and
  future.
\newblock \emph{arXiv preprint arXiv:2011.04406}, 2020.

\bibitem[He et~al.(2016)He, Zhang, Ren, and Sun]{he2016identity}
Kaiming He, Xiangyu Zhang, Shaoqing Ren, and Jian Sun.
\newblock Identity mappings in deep residual networks.
\newblock In \emph{European conference on computer vision}, pp.\  630--645.
  Springer, 2016.

\bibitem[Huang et~al.(2021)Huang, Niu, Liu, Ding, Xiao, Wu, and
  Peng]{huang2021learning}
Zhenyu Huang, Guocheng Niu, Xiao Liu, Wenbiao Ding, Xinyan Xiao, Hua Wu, and
  Xi~Peng.
\newblock Learning with noisy correspondence for cross-modal matching.
\newblock \emph{Advances in Neural Information Processing Systems},
  34:\penalty0 29406--29419, 2021.

\bibitem[Jiang et~al.(2018)Jiang, Zhou, Leung, Li, and
  Fei-Fei]{Jiang2018MentorNetLD}
Lu~Jiang, Zhengyuan Zhou, Thomas Leung, Li-Jia Li, and Li~Fei-Fei.
\newblock Mentornet: Learning data-driven curriculum for very deep neural
  networks on corrupted labels.
\newblock In \emph{ICML}, 2018.

\bibitem[Jiang et~al.(2020)Jiang, Huang, Liu, and Yang]{Jiang2020BeyondSN}
Lu~Jiang, Di~Huang, Mason Liu, and Weilong Yang.
\newblock Beyond synthetic noise: Deep learning on controlled noisy labels.
\newblock In \emph{ICML}, 2020.

\bibitem[Krizhevsky et~al.(2009)Krizhevsky, Hinton,
  et~al.]{krizhevsky2009learning}
Alex Krizhevsky, Geoffrey Hinton, et~al.
\newblock Learning multiple layers of features from tiny images.
\newblock 2009.

\bibitem[Lee et~al.(2018)Lee, He, Zhang, and Yang]{Lee2018CleanNetTL}
Kuang-Huei Lee, Xiaodong He, Lei Zhang, and Linjun Yang.
\newblock Cleannet: Transfer learning for scalable image classifier training
  with label noise.
\newblock \emph{2018 IEEE/CVF Conference on Computer Vision and Pattern
  Recognition}, pp.\  5447--5456, 2018.

\bibitem[Li et~al.(2020{\natexlab{a}})Li, Socher, and Hoi]{Li2020DivideMixLW}
Junnan Li, Richard Socher, and Steven C.~H. Hoi.
\newblock Dividemix: Learning with noisy labels as semi-supervised learning.
\newblock \emph{ICLR}, 2020{\natexlab{a}}.

\bibitem[Li et~al.(2020{\natexlab{b}})Li, Xiong, and Hoi]{li2020mopro}
Junnan Li, Caiming Xiong, and Steven~CH Hoi.
\newblock Mopro: Webly supervised learning with momentum prototypes.
\newblock \emph{arXiv preprint arXiv:2009.07995}, 2020{\natexlab{b}}.

\bibitem[Li et~al.(2020{\natexlab{c}})Li, Zhou, Xiong, and
  Hoi]{li2020prototypical}
Junnan Li, Pan Zhou, Caiming Xiong, and Steven~CH Hoi.
\newblock Prototypical contrastive learning of unsupervised representations.
\newblock \emph{arXiv preprint arXiv:2005.04966}, 2020{\natexlab{c}}.

\bibitem[Li et~al.(2021)Li, Xiong, and Hoi]{li2021learning}
Junnan Li, Caiming Xiong, and Steven~CH Hoi.
\newblock Learning from noisy data with robust representation learning.
\newblock In \emph{Proceedings of the IEEE/CVF International Conference on
  Computer Vision}, pp.\  9485--9494, 2021.

\bibitem[Li et~al.(2017{\natexlab{a}})Li, Wang, Li, Agustsson, and
  Van~Gool]{li2017webvision}
Wen Li, Limin Wang, Wei Li, Eirikur Agustsson, and Luc Van~Gool.
\newblock Webvision database: Visual learning and understanding from web data.
\newblock \emph{arXiv preprint arXiv:1708.02862}, 2017{\natexlab{a}}.

\bibitem[Li et~al.(2017{\natexlab{b}})Li, Yang, Song, Cao, Luo, and
  Li]{Li2017LearningFN}
Yuncheng Li, Jianchao Yang, Yale Song, Liangliang Cao, Jiebo Luo, and Li-Jia
  Li.
\newblock Learning from noisy labels with distillation.
\newblock \emph{2017 IEEE International Conference on Computer Vision (ICCV)},
  pp.\  1928--1936, 2017{\natexlab{b}}.

\bibitem[Liu et~al.(2020)Liu, Niles-Weed, Razavian, and
  Fernandez-Granda]{liu2020early}
Sheng Liu, Jonathan Niles-Weed, Narges Razavian, and Carlos Fernandez-Granda.
\newblock Early-learning regularization prevents memorization of noisy labels.
\newblock \emph{Advances in neural information processing systems},
  33:\penalty0 20331--20342, 2020.

\bibitem[Massey~Jr(1951)]{massey1951kolmogorov}
Frank~J Massey~Jr.
\newblock The kolmogorov-smirnov test for goodness of fit.
\newblock \emph{Journal of the American statistical Association}, 46\penalty0
  (253):\penalty0 68--78, 1951.

\bibitem[Nishi et~al.(2021)Nishi, Ding, Rich, and
  Hollerer]{nishi2021augmentation}
Kento Nishi, Yi~Ding, Alex Rich, and Tobias Hollerer.
\newblock Augmentation strategies for learning with noisy labels.
\newblock In \emph{Proceedings of the IEEE/CVF Conference on Computer Vision
  and Pattern Recognition}, pp.\  8022--8031, 2021.

\bibitem[Reed et~al.(2014)Reed, Lee, Anguelov, Szegedy, Erhan, and
  Rabinovich]{reed2014training}
Scott Reed, Honglak Lee, Dragomir Anguelov, Christian Szegedy, Dumitru Erhan,
  and Andrew Rabinovich.
\newblock Training deep neural networks on noisy labels with bootstrapping.
\newblock \emph{arXiv preprint arXiv:1412.6596}, 2014.

\bibitem[Szegedy et~al.(2017)Szegedy, Ioffe, Vanhoucke, and
  Alemi]{szegedy2017inception}
Christian Szegedy, Sergey Ioffe, Vincent Vanhoucke, and Alexander~A Alemi.
\newblock Inception-v4, inception-resnet and the impact of residual connections
  on learning.
\newblock In \emph{Thirty-first AAAI conference on artificial intelligence},
  2017.

\bibitem[Tanaka et~al.(2018)Tanaka, Ikami, Yamasaki, and
  Aizawa]{Tanaka2018JointOF}
Daiki Tanaka, Daiki Ikami, T.~Yamasaki, and Kiyoharu Aizawa.
\newblock Joint optimization framework for learning with noisy labels.
\newblock \emph{2018 IEEE/CVF Conference on Computer Vision and Pattern
  Recognition}, pp.\  5552--5560, 2018.

\bibitem[Vahdat(2017)]{Vahdat2017TowardRA}
Arash Vahdat.
\newblock Toward robustness against label noise in training deep discriminative
  neural networks.
\newblock In \emph{NIPS}, 2017.

\bibitem[Wang et~al.(2022)Wang, Xiao, Li, Feng, Niu, Chen, and
  Zhao]{wang2022pico}
Haobo Wang, Ruixuan Xiao, Yixuan Li, Lei Feng, Gang Niu, Gang Chen, and Junbo
  Zhao.
\newblock Pico: Contrastive label disambiguation for partial label learning.
\newblock \emph{arXiv preprint arXiv:2201.08984}, 2022.

\bibitem[Wang et~al.(2019)Wang, Ma, Chen, Luo, Yi, and
  Bailey]{wang2019symmetric}
Yisen Wang, Xingjun Ma, Zaiyi Chen, Yuan Luo, Jinfeng Yi, and James Bailey.
\newblock Symmetric cross entropy for robust learning with noisy labels.
\newblock In \emph{Proceedings of the IEEE/CVF International Conference on
  Computer Vision}, pp.\  322--330, 2019.

\bibitem[Wu et~al.(2021)Wu, Wei, Jiang, Mao, Tang, and Li]{wu2021ngc}
Zhi-Fan Wu, Tong Wei, Jianwen Jiang, Chaojie Mao, Mingqian Tang, and Yu-Feng
  Li.
\newblock Ngc: a unified framework for learning with open-world noisy data.
\newblock In \emph{Proceedings of the IEEE/CVF International Conference on
  Computer Vision}, pp.\  62--71, 2021.

\bibitem[Xiao et~al.(2015{\natexlab{a}})Xiao, Xia, Yang, Huang, and
  Wang]{Xiao2015LearningFM}
Tong Xiao, Tian Xia, Yi~Yang, Chang Huang, and Xiaogang Wang.
\newblock Learning from massive noisy labeled data for image classification.
\newblock \emph{2015 IEEE Conference on Computer Vision and Pattern Recognition
  (CVPR)}, pp.\  2691--2699, 2015{\natexlab{a}}.

\bibitem[Xiao et~al.(2015{\natexlab{b}})Xiao, Xia, Yang, Huang, and
  Wang]{xiao2015learning}
Tong Xiao, Tian Xia, Yi~Yang, Chang Huang, and Xiaogang Wang.
\newblock Learning from massive noisy labeled data for image classification.
\newblock In \emph{Proceedings of the IEEE conference on computer vision and
  pattern recognition}, pp.\  2691--2699, 2015{\natexlab{b}}.

\bibitem[Yi \& Wu(2019{\natexlab{a}})Yi and Wu]{Yi2019ProbabilisticEN}
Kun Yi and Jianxin Wu.
\newblock Probabilistic end-to-end noise correction for learning with noisy
  labels.
\newblock \emph{2019 IEEE/CVF Conference on Computer Vision and Pattern
  Recognition (CVPR)}, pp.\  7010--7018, 2019{\natexlab{a}}.

\bibitem[Yi \& Wu(2019{\natexlab{b}})Yi and Wu]{yi2019probabilistic}
Kun Yi and Jianxin Wu.
\newblock Probabilistic end-to-end noise correction for learning with noisy
  labels.
\newblock In \emph{Proceedings of the IEEE/CVF Conference on Computer Vision
  and Pattern Recognition}, pp.\  7017--7025, 2019{\natexlab{b}}.

\bibitem[Yu et~al.(2019)Yu, Han, Yao, Niu, Tsang, and Sugiyama]{Yu2019HowDD}
Xingrui Yu, Bo~Han, Jiangchao Yao, Gang Niu, Ivor Wai-Hung Tsang, and Masashi
  Sugiyama.
\newblock How does disagreement help generalization against label corruption?
\newblock In \emph{ICML}, 2019.

\bibitem[Zhang et~al.(2017)Zhang, Cisse, Dauphin, and
  Lopez-Paz]{zhang2017mixup}
Hongyi Zhang, Moustapha Cisse, Yann~N Dauphin, and David Lopez-Paz.
\newblock mixup: Beyond empirical risk minimization.
\newblock \emph{arXiv preprint arXiv:1710.09412}, 2017.

\bibitem[Zheltonozhskii et~al.(2021)Zheltonozhskii, Baskin, Mendelson,
  Bronstein, and Litany]{DBLP:journals/corr/abs-2103-13646}
Evgenii Zheltonozhskii, Chaim Baskin, Avi Mendelson, Alex~M. Bronstein, and
  Or~Litany.
\newblock Contrast to divide: Self-supervised pre-training for learning with
  noisy labels.
\newblock \emph{CoRR}, abs/2103.13646, 2021.
\newblock URL \url{https://arxiv.org/abs/2103.13646}.

\end{thebibliography}
\bibliographystyle{iclr2023_conference}

\appendix
\section{Appendix}
\subsection{Empirical Vicinal Risk} \label{app:evr}
We introduce the Empirical Vicinal Risk following \citet{cordeiro2022longremix}. In the semi-supervised learning based label noise learning framework, with the labeled set $\mathcal{X}$ and unlabeled set $\mathcal{U}$ from a cleaner, the DNNs are trained to minimise the empirical vicinal risk (EVR) \citep{zhang2017mixup}:
\begin{equation} \label{eq:app:evr}
\ell_{EVR} = \frac{1}{|\mathcal{X'}|} \sum_{\mathcal{X'}} \ell_{\mathcal{X'}}(p(\tilde{y'_i}|x'_i), y'_i) + \frac{\lambda^{(\mathcal{U'})}}{|\mathcal{U'}|} \sum_{\mathcal{U'}} \ell_{\mathcal{U'}}(p(\tilde{y'_i}|x'_i), y'_i),
\end{equation}
where $l_{\mathcal{X'}}$ and $l_{\mathcal{U'}}$ denote the losses for set $\mathcal{X'}$ and $\mathcal{U'}$, which are weighted by $\lambda^{(\mathcal{U'})}$. $\mathcal{X'}$ and $\mathcal{U'}$ indicate MixMatch \citep{berthelot2019mixmatch} augmented clean and noise set:
\begin{equation} 
\begin{aligned}
\mathcal{X'} = {(x'_i,y'_i):(x'_i,y'_i) \sim f(x'_i,y'_i|x_i,y_i), (x_i,y_i) \in \mathcal{X} }, \\
\mathcal{U'} = {(x'_i,y'_i):(x'_i,y'_i) \sim f(x'_i,y'_i|x_i,y_i), (x_i,y_i) \in \mathcal{U}}, 
\end{aligned}
\end{equation}
with
\begin{equation} 
\begin{aligned}
f(x'_i,y'_i|x_i,y_i)=\frac{1}{|\mathcal{X} \cup \mathcal{U}|} \sum_{\mathcal{X} \cup \mathcal{U}} \mathbb{E}_{\lambda}[\delta(x'_i= \lambda x_i+(1-\lambda)x_j, y'_i=\lambda y_i+(1-\lambda)y_j)], 
\end{aligned}
\end{equation}
where $\delta$ is a Dirac mass centered at $(x',y')$, $\lambda \sim Beta(a,a)$, and $a$ $\in$ (0, +$\inf$).

\subsection{Other Training Details} \label{app:conig}
\subsubsection{Training configurations}
In our method, we follow most of training set-up of DivideMix\citep{Li2020DivideMixLW}. We present the detailed training configures as follows:
\begin{itemize}
    \item \textbf{CIFAR-10 and CIFAR-100}. For all the experiments on CIFAR, we train our DNN model as well as class prototypes in CPC via SGD with a momentum of 0.9, a weight decay of 0.0005, and a batch size of 128. The network is trained for 450 epochs. We set the initial learning rate as 0.02, and reduce it by a factor of 10 after 225 epochs. The warm up period for the DNN is 10 epochs. The weight $\lambda^{(\mathcal{U'})}$ is set to $\{0,25,50,150\}$ as in DivideMix.
    \item \textbf{Clthing1M}. We train our DNN model as well as class prototypes in CPC via SGD with a momentum of 0.9, a weight decay of 0.001, and a batch size of 32. The model is trained for 80 epochs. The warm up period for the DNN is 1 epoch. The initial learning rate is set as 0.002 and reduced by a factor of 10 after 40 epochs. For each epoch, we sample 1000 mini-batches from the training data. The weight $\lambda^{(\mathcal{U'})}$ is set to 0.
    \item \textbf{WebVision}. We train our DNN model as well as class prototypes in CPC via SGD with a momentum of 0.9, a weight decay of 0.001, and a batch size of 32. The model is trained for 100 epochs. The warm up period for the DNN is 1 epoch. The initial learning rate is set as 0.01 and reduced by a factor of 10 after 50 epochs. For each epoch, we sample 1000 mini-batches from the training data. The weight $\lambda^{(\mathcal{U'})}$ is set to 0.
\end{itemize}

\subsubsection{Hyper-parameter Study}
In this paper, we mainly follow the tuning procedure as in DivideMix to determine the newly introduced hyper-parameters. First of all, we initialize the hyper-parameters to $e=5\%, \tau=0.5, \alpha=1$. Then,  for the large scale real world benchmark Clothing1M and WebVision, the hyper-parameter tuning is done on the validation set of Clothing1M and transferred to WebVision. For CIFAR, a small validation set with clean data is split from training data for hyper-parameter tuning. Due to the diversity of experimental set-ups, it would be an irritating task to tune hyper-parameters for each experimental set-up, respectively. Therefore, we only tune the hyper-parameters under CIFAR-100(sym80\%) and CIFAR-100(sym50\%), and transfer the hyper-parameters obtained  under CIFAR-100(sym80\%) to the noisier set-up \emph{i.e.,} CIFAR-100(sym90\%), and those obtained under CIFAR-100(sym50\%) to the less challenge set-ups \emph{i.e.,} noise ratio lower than 50\%  and all noise ratio on CIFAR-10. 

In practical, when a clean validation set is inaccessible, it would be the difficult to tune the hyper-parameters. To shed some light to the hyper-parameter set-up in these cases, we try to conclude some empirical solutions via studying the variation of performance of CPC with respect to the newly introduced hyper-parameters on different benchmarks. According to experimental results, we find that CPC is robust in the choice of hyperparameters in the range listed in Tab.\ref{tab:hyper}. Generally, $e=5\%/10\%, \tau=0.5, \alpha=0/1$ can be a good choice in most cases.

\begin{table}[]
\caption{The variation of performance of CPC with respect to the change of hyper-parameters. The classification accuracy of DNNs is reported. The best results are indicated with  boldface. }\label{tab:hyper}
\resizebox{\columnwidth}{!}{%
\begin{tabular}{l|llllllllll}
\hline
 & \multirow{2}{*}{baseline} & \multicolumn{3}{l}{CPC Warm-up epochs (e)} & \multicolumn{3}{l}{CPC threshold ($\tau$)} & \multicolumn{3}{l}{Prototypical loss weight ($\alpha$)} \\
 &  & 5\% & 10\% & 15\% & 0.5 & 0.6 & 0.7 & 0 & 0.5 & 1 \\ \hline
CIFAR-100(sym90\%) & 41.2 & 52.32 & \textbf{54.60} & 53.7 & \textbf{54.60} & 54.33 & 54.05 & \textbf{54.60} & 54.48 & 54.51 \\
WebVision & 76.3 & \textbf{79.63} & 79.32 & 79.04 & \textbf{79.63} & 79.52 & 79.36 & 79.16 & 79.44 & \textbf{79.63} \\
CIFAR-10(asym40) & 94.60 & \textbf{94.73} & 94.68 & 94.59 & \textbf{94.73} & 94.71 & 94.65 & \textbf{94.73} & 94.68 & 94.72 \\
Clothing1M & 74.73 & \textbf{75.40} & 75.04 & 74.89 & 75.08 & 75.15 & \textbf{75.40} & 75.35 & 75.28 & \textbf{75.40}
\end{tabular}%
}
\end{table}

\subsection{Discussion on the contribution of GMM cleaner to CPC} \label{app:abl}
  
  In typical prototypical contrastive objective, the unsupervised training labels are determined by similarity between samples and prototypes. Compared to it, we empirically find that GMM cleaner provides more accurate training labels for prototypes, especially in the early stage of training. For example, in CIFAR-10(asym-40\%), the averaged accuracy of training labels from GMM cleaner is 9.7\% higher during the CPC warming up period. 
  
  To evaluate the contribution of GMM cleaner in our framework, we further present ablation study results in Tab.~\ref{tab:abl}. For \emph{CPC w/o GMM Cleaner}, we remove the GMM cleaner and learn class prototypes in CPC with prototypical contrastive objective as in MoPro \citep{li2020mopro}. In experiments, we find that without the help of the GMM cleaner, the learnt prototypes generate less accurate data partition that further drawing back the overall training framework for DNNs as shwon in Tab.~\ref{tab:abl}. The situation is especially severe on the challenging benchmark with more diverse data, \emph{e.g.,} WebVision.  The results demonstrate the benefits of the GMM cleaner in our method. 
  
  To prove the superiority of our method, we also compare the quality of prototypes learnt in our method with prototypes learnt in MoPro \citep{li2020mopro} on the first 50 classes of WebVision.
  To evaluate the quality of prototypes learnt in CPC, we utilize the prototypes to classify test samples via measuring the similarity between samples and prototypes. We implement the experiment with the official code released by the MoPro team. The results show that our prototype achieves a top1 accuracy of 78.44\%, while MoPro's accuracy is 72.23\%. The result demonstrates that our method is able to learn better prototypes.
  
\begin{table}[]
\caption{Ablation study on the contribution of GMM cleaner.The classification accuracy of DNNs is reported. The best results are indicated with  boldface.}\label{tab:abl}
\resizebox{\columnwidth}{!}{%
\begin{tabular}{l|llll}
method & CIFAR-100(sym90\%) & WebVision & CIFAR-10(asym40\%) & Clothing1M \\ \hline
Baseline & 41.2 & 76.3 & 94.6 & 74.73 \\
CPC w/o GMM Cleaner & 42.9 & 26.8 & 93.92 & 74.09 \\
CPC & \textbf{54.6} & \textbf{79.63} & \textbf{94.73} & \textbf{75.4}
\end{tabular}%
}
\end{table}

\subsection{Supplementary discussion on the theoretical justification}
\label{app:kl}
\subsubsection{Is $q(z'_{i})$ a proper approximation to $q(z_i)$ in practical?} 
\begin{figure}[t]
  \centering
  \includegraphics[width=1\columnwidth]{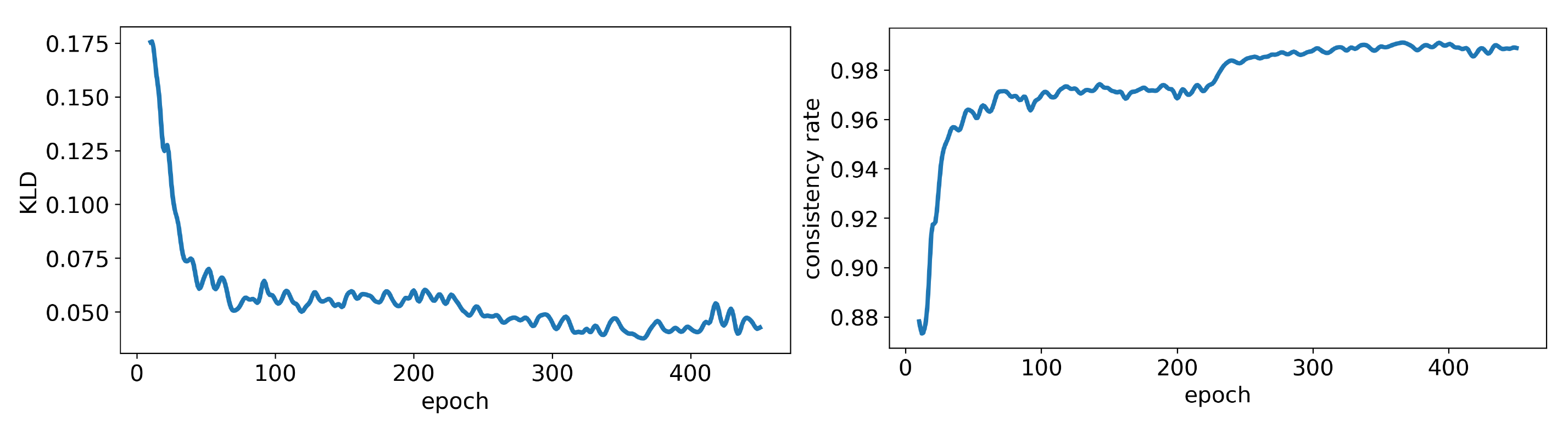}
    \vspace{-2\baselineskip}
  \caption{ The left figure shows the KLD between $q(z'_{i})$ and $q(z_i)$. The right figure presents the consistency rate between $z'_i$ and $z_i$. Results are collected from CIFAR-10-aysm40\%. }
   \label{fig:kl}
\end{figure}

In Section~\ref{sec:em}, we replace the estimation of CPC $q(z_i)$ in Eq.~(\ref{mstep}) with the estimation of GMM cleaner $q(z'_i)$ and justify $q(z'_{i})$ can be considered as an approximate to $q(z_i)$. To investigate if the approximation holds in practical, we calculate the K-L Divergence as well as classification consistency between $q(z'_{i})$ and $q(z_i)$. As shown in Figure~\ref{fig:kl}, as the training going on, the KLD  between $q(z'_{i})$ and $q(z_i)$ is converged and the classification consistency increases. 

\subsubsection{Training prototypes with \texttt{$L_C$} is an approximation to the M-step in EM} \label{app:loss}
As illustrated in Section \ref{sec:em},  in order to introduce the ``small-loss prior'' to provide stronger and more robust supervision signals to the learning of CPC, in the \textbf{E-step}, we estimate the probability distribution of clean or unclean of samples, denoted as $q(z'_i)$, via the GMM cleaner, which is an approximation to the $q(z_i)$ in Eq.~(\ref{estep}). And consequently, we replace the $q(z_i)$ in  Eq.~(\ref{mstep}) with $q(z'_i)$ and find the prototype $C$ to minimize the bound, which makes the loss function $L_C$ in Eq.~(\ref{fullloss}) an approximation to Eq.~(\ref{mstep}). The detailed analysis on the relationship between Eq.~(\ref{fullloss}) and Eq.~(\ref{mstep}) is as follows.

Firstly, we replace the estimation of CPC $q(z_i)$ in Eq.~(\ref{mstep}) with the estimation of GMM cleaner $q(z'_i)$ which is a justified approximate to $q(z_i)$:
\begin{equation} 
\begin{aligned} \label{app-mstep}
        C^{new} &=  \mathop{\arg\min}_{C} -   \sum_{ D} \sum_{z_i \in \{0,1\}} q(z_i) \log   p(y_i|C,x_i) \\
         & \approx \mathop{\arg\min}_{C} -   \sum_{ D} \sum_{z'_i \in \{0,1\}} q(z'_i) \log   p(y_i|C,x_i) \\
         &= \mathop{\arg\min}_{C} -   \sum_{ D} \left [ q(z'_i = 0) \log p(y_i|C,x_i) + q(z'_i = 1)\log p(y_i|C,x_i) \right] \\
\end{aligned}  
\end{equation}
 In Eq.~(\ref{fullloss}), $q(z'_i)$ is quantified to 1 and 0 by the threshold $\tau$, which makes it a ``hard" version to  Eq.~(\ref{app-mstep}). Specifically, the first term in Eq.~(\ref{app-mstep}) updates the prototypes $C$ to better align the samples, that classified as clean, with labeled class prototypes. It is equivalent with the effect of Eq.~(\ref{fullloss}) to positive samples, where:
 \begin{equation}
 l = \log(\mathrm{sigmoid}(v'_i c_k^\top)), k=y_i, z'_i=0
\end{equation}
 where $v'_i$ is the embedding of sample $x_i$.
 The second term in Eq.~(\ref{app-mstep}) updates $C$ to prevent the samples, that classified as noise, aligning with labeled class prototypes so as to better recognize the sample as noise (\emph{i.e., $z'_i=1$}), which is equivalent with the effect of Eq.~(\ref{fullloss}) reducing the probability of negative samples to be recognized as clean:
 \begin{equation}
 l = \log(1-\mathrm{sigmoid}(v'_i c_k^\top)), k=y_i, z'_i=1
\end{equation}

\subsection{Illustration to the overall framework} \label{app:alg}
In this paper, we plug CPC to the popular DivideMix framework. We delineate the overall training framework in Alg.\ref{alg1}.

\begin{algorithm}
	\renewcommand{\algorithmicrequire}{\textbf{Input:}}
	\renewcommand{\algorithmicensure}{\textbf{Output:}}
	\caption{CPC based DivideMix}
	\label{alg1}
	\begin{algorithmic}[1]
		\STATE \textbf{Input}: Dataset $D=(X,Y)$, DNNs $\theta^{(1)}$, $\theta^{(2)}$, CPC with class prototypes $C^{(1)}$, $C^{(2)}$, clean probability $\tau$, CPC warm-up period $e$.
		\STATE  $\theta^{(1)}, \theta^{(2)} =$ WarmUp$(X,Y,\theta^{(1)})$, WarmUp$(X,Y, \theta^{(2)})~~~$      \emph{//standard training to warm-up DNNs}
		\WHILE{ $epoch <$ MaxEpoch}
	\STATE \emph{// get GMM cleaners by loss distribution modeling and calculate clean/noise probability distribution }
		\STATE $Q^{(2)}(Z')=$GMM$(X,Y,\theta^{(1)})$
		\STATE $Q^{(1)}(Z')=$GMM$(X,Y,\theta^{(2)})$
	\STATE \emph{// calculate clean/noise probability distribution via CPC}
		\STATE $Q^{(2)}(Z)=$CPC$(X,Y,\theta^{(1)},C^{(1)})$
		\STATE $Q^{(1)}(Z)=$CPC$(X,Y,\theta^{(2)},C^{(2)})$
		\FOR{r $\in \{ 1,2 \}$} 
		\STATE \emph{// stage1 begin}
		\STATE $\mathcal{X}^{GMM(r)}=\{(x_i,y_i,w_i)|w_i=q^{(r)}(z'_i=0), q^{(r)}(z'_i=0) > \tau, (x_i,y_i) \in D, q^{(r)}(z'_i=0) \in Q^{(r)}(Z'=0) \}$
		\STATE $\mathcal{U}^{GMM(r)} = \{x_i| q^{(r)}(z'_i=0) \leq \tau, x_i \in X, q^{(r)}(z'_i=0) \in Q^{(r)}(Z'=0)\} $
		\STATE Get noise labels $\{y_i|(x_i,y_i) \in D, x_i \in \mathcal{U}^{GMM(r)}\}$
		\STATE Update $C^{k}$ based on Eq.\ref{fullloss}
		\STATE \emph{// stage1 end}
		\STATE \emph{// stage2 begin}
	    \IF{$epoch < e$}
        \STATE $\mathcal{X}^{(r)}=\mathcal{X}^{GMM(r)}$, $\mathcal{U}^{(r)}=\mathcal{U}^{GMM(r)}$ \emph{$~~~~~$//use data partition from GMM cleaner to update DNNs during the CPC warm-up period}
        \ELSE
        \STATE $\mathcal{X}^{(r)}=\{(x_i,y_i,w_i)|w_i=q^{(r)}(z_i=0), q^{(r)}(z_i=0) > \tau, (x_i,y_i) \in D , q^{(r)}(z_i=0) \in Q^{(r)}(Z=0)\}$
		\STATE $\mathcal{U}^{(r)} = \{x_i|q^{(r)}(z_i=0) \leq \tau, x_i \in X, q^{(r)}(z_i=0) \in Q^{(r)}(Z=0)\} $
	    \ENDIF
	    \STATE Update $\theta^{r}$ based on Eq.\ref{eq:app:evr} as in standard DivideMix
	    \STATE \emph{// stage2 end}
		\ENDFOR
		\STATE $epoch \leftarrow epoch + 1$
		\ENDWHILE
		\ENSURE  DNNs $\theta^{(1)}$,  $\theta^{(2)}$
	\end{algorithmic}  
\end{algorithm}

\end{document}